\documentclass{article}

\usepackage[margin=1in]{geometry}

\usepackage{amsmath, amssymb}

\usepackage{graphicx}
\usepackage{wrapfig}    

\usepackage{hyperref}

\usepackage{tcolorbox}
\usepackage{booktabs}
\usepackage{pgfplots}

\usepackage[utf8]{inputenc}
\usepackage{caption}   
\usepackage{subcaption} 

\usepackage{colortbl}

\pgfplotsset{compat=1.17}
\title{Towards More Effective Table-to-Text Generation: Assessing In-Context Learning and Self-Evaluation with Open-Source Models}

  \author{
    Sahar Iravani, {Zuse  Institute Berlin, iravani@zib.de} \and 
    Tim O. F. Conrad, {Zuse  Institute Berlin, conrad@zib.de}
}      
\date{}

\begin{document}

\maketitle

\begin{abstract}
Table processing, a key task in natural language processing, has significantly benefited from recent advancements in language models (LMs). However, the capabilities of LMs in table-to-text generation, which transforms structured data into coherent narrative text, require an in-depth investigation, especially with current open-source models. This study explores the effectiveness of various in-context learning strategies in LMs across benchmark datasets, focusing on the impact of providing examples to the model. More importantly, we examine a real-world use case, offering valuable insights into practical applications. To complement traditional evaluation metrics, we employ a large language model (LLM) self-evaluation approach using chain-of-thought reasoning and assess its correlation with human-aligned metrics like BERTScore. 
Our findings highlight the significant impact of examples in improving table-to-text generation and suggest that, while LLM self-evaluation has potential, its current alignment with human judgment could be enhanced. This points to the need for more reliable evaluation methods.

\end{abstract}

\section{Introduction}
\begin{wrapfigure}{r}{0.38\textwidth} 
    \centering
    \includegraphics[width=0.38\textwidth]{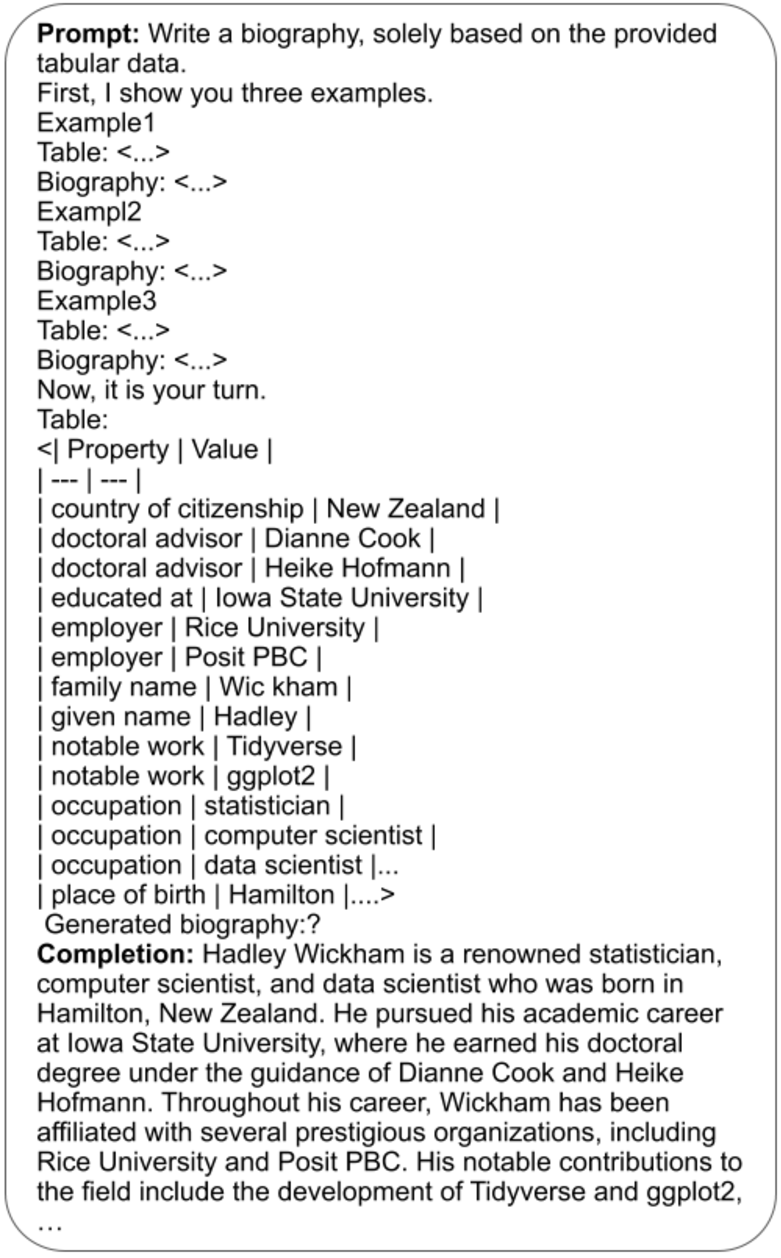}
    \caption{An example of in-context learning for table-to-text 
    \label{fig:simple-example}generation.}
\end{wrapfigure}

In today's data-driven world, the ability to make informed decisions increasingly depends on our capacity to process and interpret structured data. As we engage with diverse forms of such data across various domains, there is a growing demand for methods that can transform complex, structured information into clear and accessible content \cite{Fang2024, li2024sheetcopilot, lu2024large}. An area that has received significant attention is table processing \cite{chen2020wikitablet, chen2022large, hadi2023survey}, driven by advances in language models (LMs), which have revolutionized numerous natural language processing (NLP) tasks \cite{ram2023context, hadi2023survey, brown2020language, thirunavukarasu2023large}. These models, trained on vast amounts of text, excel at recognizing patterns in language and applying them to a variety of applications. Among the most promising techniques for handling table-related tasks is the fine-tuning of pre-trained LMs on datasets designed for specific tasks \cite{chen2020logical, liu2022plog, zhao2023loft}. However, identifying or developing a dataset that is appropriately tailored to the capabilities and requirements of LMs can be challenging. In response, researchers have explored in-context learning \cite{brown2020language}, a strategy where models are guided by tailored prompts and minimal examples, improving performance across a wide range of NLP challenges. This approach has been particularly effective in table-based tasks like question answering and fact-checking \cite{chen2022large, nan2023enhancing}. Building on this, chain-of-thought prompting methods \cite{wang2022self,wei2022chain} further empower LMs to perform complex reasoning tasks, enabling them to tackle challenges that require multi-step logic and in-depth understanding \cite{han2022folio, zhao2023investigating, ye2023large}.

Despite these advancements, the potential of recent open-source LMs in generating narrative text from tabular data or the table-to-text generation task remains underexplored.
This task focuses in particular on extracting and verbalizing insights from tabular data into narrative text. Automating this process has far-reaching implications for fields such as healthcare, business intelligence, and academic research, especially in automated report generation and personalized data summaries. It minimizes manual effort, ensures real-time updates at lower costs, and enables more informed decision-making.

\subsection{Investigating In-Context Learning for Table-to-Text Generation}

In this study, our objective is to comprehensively explore the performance of in-context learning in the table-to-text generation task. An example of this task is presented in Figure \ref{fig:simple-example}. We investigate a real-world use-case scenario of generating biographies for mathematicians. For this use-case, structured (tabular) data is available from sources such as WikiData or the \textit{Mathematical Research Data Initiative} consortium (MaRDI) which focuses on systematic management and utilization of mathematical research data \cite{conrad2023making}.

We focus on optimizing prompting strategies, including zero-shot, single-shot, and few-shot approaches, to enhance table-to-text generation. By carefully selecting examples, we aim to improve model performance, leveraging recent advancements in in-context learning \cite{chen2022large}. To evaluate the presented approach, we run experiments on two benchmark datasets: WikiBio \cite{DBLP:journals/corr/LebretGA16} and ToTTo \cite{parikh2020totto}. Here, table-to text generation is investigated in two scenarios: WikiBio, where the text is the description of a table, and ToTTo, where a sentence description is presented for a specific cell of the table. We also consider the concern that LMs may have been exposed to portions of public datasets, such as Wikipedia-based datasets, during training, which could inadvertently bias their performance and lead to inflated results. To address this, we evaluate the models on recent Wikipedia biography pages created after the public release of the language models used in this study. This approach mitigates the risk of data overlap, providing a more accurate reflection of the models' unbiased capabilities in transforming structured data into narrative text.


\begin{figure}[t] 
    \centering
    \includegraphics[width=\textwidth]{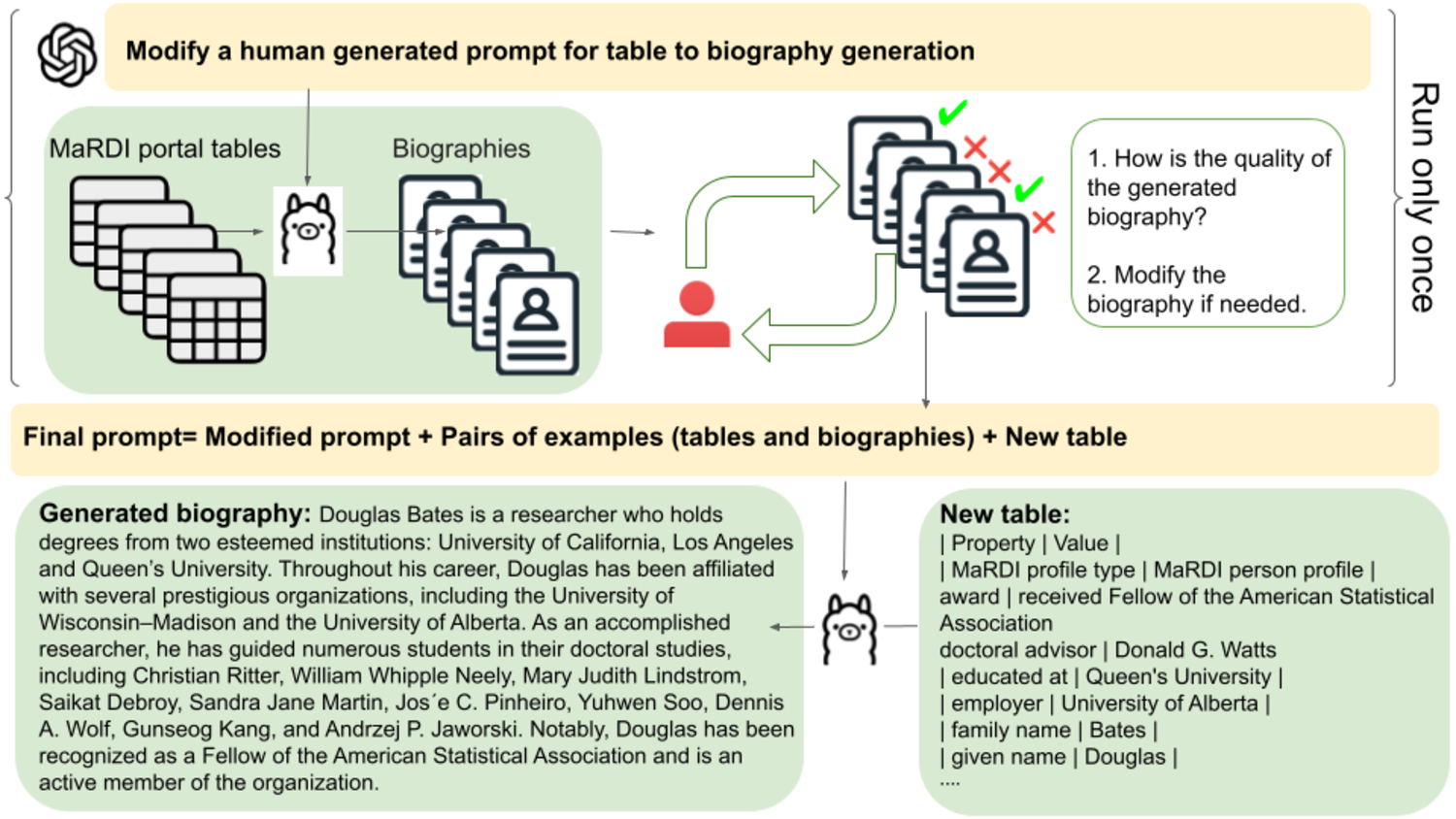} 
    \caption{The real-world table-to-text generation scenario investigated in this paper.}
    \label{fig:Illustration}
\end{figure}



In addition to these datasets, we explore a specialized use case involving the MaRDI Portal \cite{conrad2023making} (as illustrated in Figure \ref{fig:Illustration}), which contains structured information on mathematicians and their contributions. This investigation provides a unique opportunity to apply table-to-text generation in a real-world scenario. The MaRDI use case is particularly challenging because the available (structured) data about a person can be incomplete, due to the way data are collected from various sources. We present how the choice of examples in the few-shot approach can lead to fewer hallucinations in a real-world table-to-text task, biography generation from tabular data. This comprehensive study allows us to better understand the strengths and limitations of LMs in diverse and complex environments, ultimately contributing to more effective and reliable table-to-text generation applications. Although many high-performing open-source language models exist, in this work we focus on two models: Llama 3 \cite{dubey2024llama} and Phi-3 \cite{abdin2024phi} which are good representative of the current landscape and allow us to allocate resources for comprehensive evaluations, including various in-context learning strategies and real-world use cases. Llama 3, as one of the largest open-source and most advanced models, provides insights on large-scale architectures in table-to-text generation. In contrast, Phi-3, a more lightweight model, allows us to examine performance relative to model size and efficiency. This combination allows us to explore a spectrum of capabilities and offers a balanced perspective on the effectiveness of model scales in practical applications.

\section{Methodology}
In this study, our objective was to explore the effectiveness of in-context learning in table-to-text generation tasks with current open-source capable language models including Llama 3 and Phi-3, particularly focusing on the performance differences between zero-shot, single-shot, and few-shot prompting strategies. Our investigation was conducted through two distinct table-to-text generation tasks: (1) generating a concise one-line description from a specific table and (2) generating a comprehensive biography from a given table.

\paragraph{In-context Learning:} To investigate the impact of in-context learning, we employed three approaches: zero-shot, single-shot, and few-shot prompting. For the zero-shot prompting, we provided the model with a simple instruction to generate text from the table without any examples. This approach allowed us to assess the model’s ability to perform the task without prior examples. In the single-shot approach, we added one example input and corresponding output before generating the text for the target table. In the few-shot approach, we supplied the model with three sets of examples. These approaches were used to evaluate how providing contextual examples influenced the model's performance in generating accurate and contextually relevant text.


Inspired by the study by Zhou et al. \cite{zhou2022large}, we generated the initial prompts for our experiments using the well-established GPT-4 system. For each task, we began by providing GPT-4 with a simple task description and a single example from the dataset to generate initial prompts for other language models. However, the output from GPT-4 was not used without modification in all cases. The complexity of the task and the specific characteristics of the dataset often necessitated iterations on the generated prompts. We refined the prompts to ensure that the results were more closely aligned with our desired results. An example of this refinement within our MaRDI use case is as follows: one row in the dataset contains gender-related information, such as [ sex or gender \texttt{|} male ] (refer to Figure \ref{fig:mardi-dataset-example}). Our initial experiment with the GPT4 generated prompt, resulted in generated text where the gender was explicitly linked to the scientist’s name, such as: \textit{Douglas Bate is a \textbf{male} researcher.}, which was not aligned with our objectives. Since we opted not to preprocess the data to remove such patterns, we instead added a refinement to the prompt: \textit{Generate the biography without directly mentioning the gender}. This adjustment mitigated the issue without altering the dataset.
Additionally, in the cases of single-shot and few-shot prompting, we manually selected examples from the ground truth samples to further guide the models. After evaluating the model's performance across different samples in a zero-shot setting, we selected one high-quality example where the model performed well to serve as the prompt for single-shot learning. For the few-shot setup, we supplemented this optimal example with two additional examples where the model demonstrated moderate performance in the zero-shot scenario, as well as one example where the model performed poorly. This selection strategy ensured that a diverse range of samples was represented, allowing the model to generalize better across varying input complexities.

We integrated these examples with natural language instructions. Since LMs process input in a linear, sequential text format, a common method, as highlighted in previous research \cite{singha2023tabular, sui2024table}, is to linearize tables into a markdown structure. This is achieved by separating rows with new lines and using column separators (e.g., \texttt{"|"}
) between individual cells to maintain the table's structure within the text sequence.


\subsection{Experimental Setup}
The experiments were conducted on a Linux based system having 16 CPU cores, 64 GB RAM, and a NVIDIA A100 GPU for model inference tasks. Language models were deployed using the Ollama framework (version 0.1.44) for model management and inference. We used Python 3.11 with the Langchain module (langchain$\_$community), enabling seamless loading and interaction with Ollama.
We conducted experiments using two recent open-source language models, Llama 3 \cite{dubey2024llama} with 70B parameters as a large language model (LLM) and Phi-3 \cite{abdin2024phi} with 14B parameters as a small language model (SLM). This selection enabled a robust evaluation across different scales. LLaMA 3, developed by Meta AI, is known for its high optimization and performance across diverse NLP tasks. It employs a unique 128K tokenizer and grouped query attention to enhance inference speeds. The model is trained on 15 trillion tokens, emphasizing multilingual capabilities and advanced attention mechanisms.
Microsoft's Phi-3 is a compact language model optimized for both efficiency and high performance. Renowned for employing high-quality training datasets, Phi-3 utilizes a curated dataset that includes heavily filtered web data, curated educational materials, and synthetic data created by larger models. Despite its reduced parameter count, this model consistently surpasses larger counterparts in standard benchmarks, demonstrating its superior design and training methodology.

\subsection{Measurements}
\paragraph{BLEU} \cite{papineni2002bleu} 
(Bilingual Evaluation Understudy) score evaluates the 
quality of machine-generated text by comparing them to one or more reference text. It works by calculating the overlap between small sequences of words, called n-grams, from the generated text and the reference texts. BLEU considers multiple n-gram lengths (e.g., single words, pairs of words, etc.) to capture different levels of similarity. To avoid giving too much credit for repeated words, BLEU uses a method called clipping, which limits how often n-grams are counted. Additionally, a brevity penalty is applied to gnerated text that are too short. The final BLEU score is the geometric mean of the n-gram precisions, weighted equally, and scaled by the brevity penalty.
The BLEU score was initially introduced for the evaluation of machine translation but has since been widely adopted across various natural language processing (NLP) tasks. However, it exhibits limitations in effectively capturing semantic nuances and may not consistently correlate with human judgment, particularly in more complex NLP tasks \cite{reiter2018structured, ji2023survey}.
\paragraph{BERTScore} \cite{zhang2019bertscore} evaluates the semantic similarity between a candidate and reference sentence using token embeddings from pre-trained models like BERT. Unlike BLEU, BERTScore captures semantic meaning rather than exact word matches, providing a more nuanced assessment of text quality. To calculate BERTScore, both sentences are tokenized and embedded, and then cosine similarity is computed between each token in the candidate and every token in the reference. Precision is calculated by averaging the maximum similarity for each candidate token with respect to the reference, while recall averages the maximum similarity for each reference token with respect to the candidate. The final BERTScore is the harmonic mean (F1-score) of precision and recall.


\paragraph{LLM Self Evaluation} employs LLMs to evaluate the quality of a text based on a set of pre-defined criteria.
We tested a LLM self-evaluation strategy inspired by the G-EVAL framework \cite{liu2023g}, which incorporates LLMs with chain-of-thought (CoT) reasoning \cite{wei2022chain}. We specifically tailored the evaluation template, originally developed for summarization tasks, to suit biography evaluation. G-Eval employed GPT-4 to generate CoT prompts and evaluate generated text based on four key criteria: fluency, relevance, consistency, and coherence. In this study, we instead employed Llama 3 to assess the generated biographies based on the GPT-4 generated criteria.
We were particularly interested in determining whether LLM self-evaluation could be effectively used for table-to-text generation, especially for datasets like MaRDI that lack reference data to measure performance. Our goal was to assess the reliability of this evaluation method in scenarios where traditional metrics could not be applied, ensuring a comprehensive assessment even in the absence of benchmark data.


\subsection{Datasets}
\subsubsection{WikiBio}
The WikiBio dataset \cite{lebret2016neural} consists of approximately 700,000 pairs of Wikipedia infoboxes and their corresponding biographical introductions. Each infobox is a structured collection of key-value pairs representing factual information (e.g., name, birth date, occupation), while the corresponding biography provides a natural language summary of these facts. The dataset is widely used in natural language generation tasks, specifically for table-to-text generation. WikiBio presents challenges such as maintaining factual fidelity, ensuring text fluency, and handling varying levels of detail across infoboxes, making it a suitable resource for evaluating table-to-text generation models. An example of this dataset is illustrated in Figure \ref{fig:wikibio-dataset-example}.

        
        
        

\subsubsection{Recent Wikipedia pages}

To ensure a fair evaluation of the latest language models, Llama 3 and Phi-3, we specifically designed a dataset similar to WikiBio, but with a critical distinction: it exclusively comprised Wikipedia biography pages created after June 2024. This selection criterion was pivotal because it coincides with the timeline after the public releases of Llama 3 in April 2024 and Phi-3 in May 2024. Consequently, this approach decreased the likelihood that the models had previously encountered the data during training (unless the recent Wikipedia pages were generated by these models themselves). By focusing on these newly generated pages, we aimed to provide a more accurate and unbiased assessment of the models' capabilities in executing table-to-text generation tasks.

\subsubsection{MaRDI} \label{subsubsection:MaRDI}

In our use case, we focused on generating biographical content from structured mathematical research data housed within the MaRDI Portal \cite{conrad2023making}. An example of this dataset is illustrated in Figure \ref{fig:mardi-dataset-example}. This portal featured a rich knowledge graph that incorporated data from various mathematical sources such as DLMF, CRAN, PolyDB, swMATH, and zbMATH, which were still in the partial stages of integration. The objective was to automate the generation of detailed biographies for mathematicians and researchers by harnessing this structured data, thereby facilitating a more dynamic presentation of their contributions and impact within the mathematical community. As we do not have any reference biographies for this data, we evaluated the outcomes manually and with the LLM self-evaluation method across relevance, fluency, consistency, and coherence metrics.

For this dataset, we implemented the zero-shot prompting strategy, similar to our approach with other datasets. However, since we lacked existing examples for single-shot and few-shot prompting, we adopted this strategy: first, we generated initial outputs using the zero-shot method. We then manually modified selected samples. For the single-shot prompting, we used one example of a well-constructed biography to provide a clear standard for the model to follow when generating new outputs. Our experiments revealed that most hallucinations occurred in cases where the information was incomplete or missing. Therefore, to select the most effective examples for few-shot prompting, we included both the samples where the model performed well, to serve as a template for replicating success in other cases, and those samples that contained significant errors, requiring substantial editing to serve as an example. 
 By including both successful outputs and those with significant errors, we aimed to guide the model on how to handle such situations and cover variety in the data.


\subsubsection{ToTTo}

\begin{figure}[htbp]
    \centering
    \begin{minipage}[t]{0.54\textwidth}
        \centering
        \includegraphics[width=\textwidth]{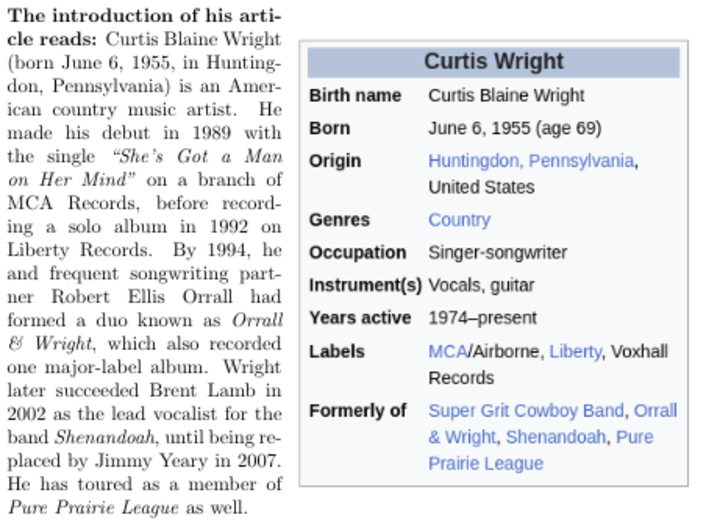}
        \caption{An example in the Wikibio dataset: the pairs of infobox and biographical introductions. The goal of the task is given the infobox, generate a biography style description.
        The biography generated by our pipeline is:  \textit{“Curtis Wright (born June 6, 1955, in Huntingdon, Pennsylvania, United States) is a country singer-songwriter. He has been active in the music industry since 1989 and is known for his solo performances as well as his work with various associated acts such as Vern Gosdin, Robert Ellis Orrall, Shenandoah, and others. Wright's music career has spanned multiple decades, during which he has released albums under labels including MCA/Airborne and has been part of groups like Orrall \& Wright, Free Falls, Cumberland Road, Liberty, and Giant.”}
 \cite{lebret2016neural}.}
        \label{fig:wikibio-dataset-example}
    \end{minipage}
    \hspace{0.05\textwidth} 
    \begin{minipage}{0.39\textwidth}
        \centering
        \includegraphics[width=0.7\textwidth]{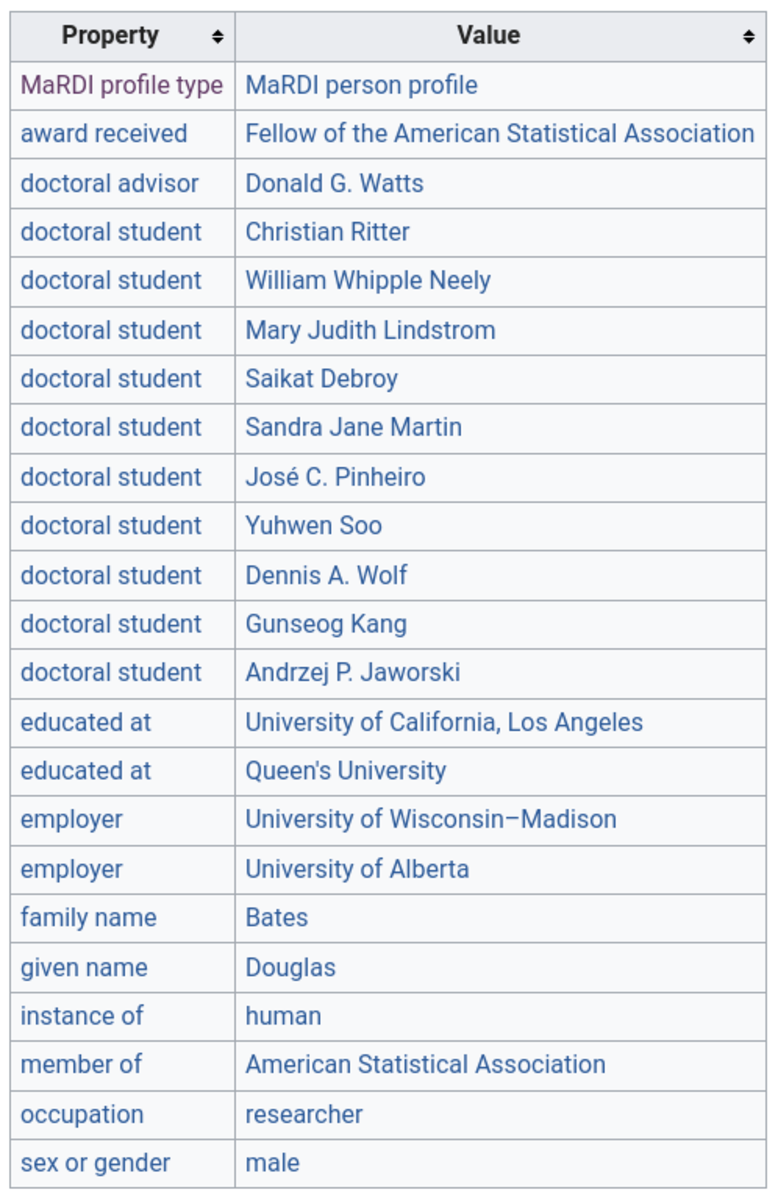}
        \caption{An example in the Mardi dataset. The biography generated by our pipeline is: “\textit{Douglas Bates is a researcher who holds degrees from two esteemed institutions: University of California, Los Angeles and Queen's University. Throughout his career, Douglas has been affiliated with several prestigious organizations, including the University of Wisconsin–Madison and the University of Alberta. As an accomplished researcher, he has guided numerous students in their doctoral studies, including Christian Ritter, William Whipple Neely, Mary Judith Lindstrom, Saikat Debroy, Sandra Jane Martin, José C. Pinheiro, Yuhwen Soo, Dennis A. Wolf, Gunseog Kang, and Andrzej P. Jaworski. Notably, Douglas has been recognized as a Fellow of the American Statistical Association and is an active member of the organization.}”
}
        \label{fig:mardi-dataset-example}
    \end{minipage}
    
    \vspace{1em} 
    
    \begin{minipage}{\textwidth}
        \centering
        \includegraphics[width=0.8\textwidth]{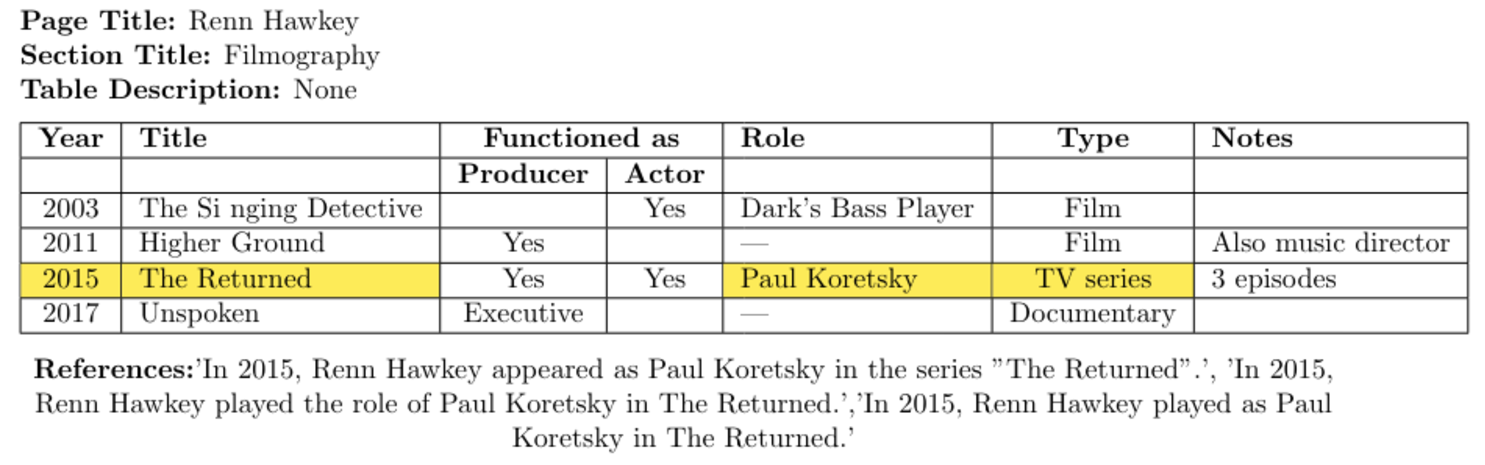}
        \caption{Example in the ToTTo dataset: a pair of table with highlighted cells and hand-crafted reference descriptions. The goal of the task is given the table, table metadata (such as the title), and set of highlighted cells, to produce a description. \cite{parikh2020totto}. The description generated by our pipeline is: “\textit{In the 2015 TV series “The Returned”, Renn Hawkey played the role of Paul Koretsky.}”}
        \label{fig:totto-dataset-example}
    \end{minipage}
\end{figure}

The ToTTo dataset \cite{parikh2020totto} is a large-scale corpus for table-to-text generation, containing over 120,000 examples of Wikipedia tables paired with human-written descriptions. It focuses on generating fluent and factually accurate natural language summaries from specific subsets of tables, where relevant cells are annotated to guide the content selection process. ToTTo addresses key challenges such as content selection, fluency, and factual accuracy, making it a critical benchmark for models designed to convert structured table data into coherent text. Our aim in running the experiment on this task was to observe how well LMs could understand where to focus within the table to produce accurate and relevant descriptions.  An example of this dataset is illustrated in Figure \ref{fig:totto-dataset-example}.

\section{Results}

In this section, we present the experimental results, benchmarking the performance of our pipeline across multiple datasets. Figures  \ref{fig:wikibio-dataset-example},\ref{fig:mardi-dataset-example}, and \ref{fig:totto-dataset-example} provide illustrative examples of the pipeline’s output, demonstrating its effectiveness across diverse datasets. Additionally, Figure \ref{fig:mardi_0-1-few-shot-comparision} highlights the impact of example selection in few-shot prompting, showcasing how carefully chosen examples enhance the relevance of the generated outputs. In the following subsections, we provide a detailed walkthrough of the benchmarking results, underscoring the robustness and versatility of the pipeline in benchmark datasets and a real-world application.

\subsection{WikiBio}

Table \ref{tab:WikiBio_0-1-few-shot-comparision} presents the comparison study of the zero-shot, single-shot and few-shot prompting on the WikiBio dataset. 
BLEU scores and BERTScores show incremental performance improvement with Llama 3 by employing examples in the prompt.
However, Phi-3, as a SLM, exhibited comparatively less benefit. We also conducted a manual evaluation of 10 randomly selected samples. While Llama 3 demonstrated slight improvements and a reduction in hallucinations, no enhancement in the quality of biographies was observed when examples were provided to the Phi-3 model.
To ensure that this difference in performance was not due to Phi-3 having already been exposed to the data during its training (which could diminish the impact of additional examples) we conducted experiments on recently added Wikipedia pages.
The LLM self-evaluation results in Table \ref{tab:WikiBio_0-1-few-shot-comparision} across fluency, relevance, consistency, and coherence did not show any agreement with the quantitative measurements. In many cases, the self-evaluation scores were close to the maximum possible values, providing little meaningful differentiation between the different prompting strategies. We were more interested to see if these metrics had any agreement with BERTScores that have been shown to correlate well with human judgment. In Figure \ref{fig:wiki-heatmap} the correlation heatmap of all metrics are illustrated. As evident, the self-evaluation metrics do not show any agreement with Bertscores.
We further examined this discrepancy using recent Wikipedia pages to gain a better understanding of its applicability in table-to-text generation tasks.

\begin{table}[b]
\centering
\caption{Comparison study of different prompting strategies on the WikiBio dataset.}
\label{tab:WikiBio_0-1-few-shot-comparision}
\begin{tabular}{lccc|ccc}
\toprule
& \multicolumn{3}{c|}{Llama 3} & \multicolumn{3}{c}{Phi-3} \\

Metrics                        & Zero Shot & Single Shot & Few Shot & Zero Shot & Single Shot & Few Shot \\
\midrule
Relevance                & \textbf{4.996} & 4.984    & 4.982 & \textbf{4.917}  & 4.702    & 4.649 \\
Fluency                  & 2.926      & \textbf{2.972} & 2.968    & 2.469  & 2.432    & \textbf{2.603} \\
Consistency               & \textbf{4.998} & 4.990    & 4.982 & \textbf{4.817}  & 4.568    & 4.534 \\
Coherence                 & 4.961   & \textbf{4.970} & 4.959 & \textbf{4.945}  & 4.622    & 4.728 \\
\hline
BLEU                    & 0.401   & 0.411    & \textbf{0.422} & \textbf{0.470}  & 0.407    & 0.356 \\
BLEU2                     & 0.381   & 0.391    & \textbf{0.408} & \textbf{0.460}  & 0.397    & 0.346 \\
BLEU3                     & 0.312   & 0.322    & \textbf{0.341} & 0.230  & \textbf{0.371}    & 0.287 \\
BLEU4                     & 0.256   & 0.267    & \textbf{0.285} & \textbf{0.348}  & 0.309    & 0.235 \\
BERTScore Precision           & 0.543   & 0.557    & \textbf{0.577} & 0.533           & \textbf{0.557}    & 0.523 \\
BERTScore Recall              & 0.622   & 0.651    & \textbf{0.657} & 0.632           & \textbf{0.641}    & 0.638 \\
BERTScore Mean                & 0.573   & 0.592    & \textbf{0.606} & \textbf{0.571}  & 0.587    & 0.566 \\
\bottomrule
\end{tabular}
\end{table}

\subsection{Recent Wikipedia pages}

Table \ref{tab:recent_wiki_0-1-few-shot-comparision} presents the comparison study of the zero-shot, single-shot, and few-shot prompting on recent Wikipedia pages.
According to this table, the few-shot approach generally outperforms zero-shot and single-shot approaches across most metrics in both Llama 3 and Phi-3 models. This suggests that incorporating a few examples into the model's training or inference process could enhance its performance. The performance improvements are notably more pronounced with Llama 3 compared to Phi-3. By manually comparing the 10 randomly selected samples we also noticed more considerable improvement with Llama 3 than Phi-3.

\begin{figure}[h]
    \centering
    \begin{subfigure}[b]{0.45\textwidth}
        \centering
        \includegraphics[width=0.8\textwidth]{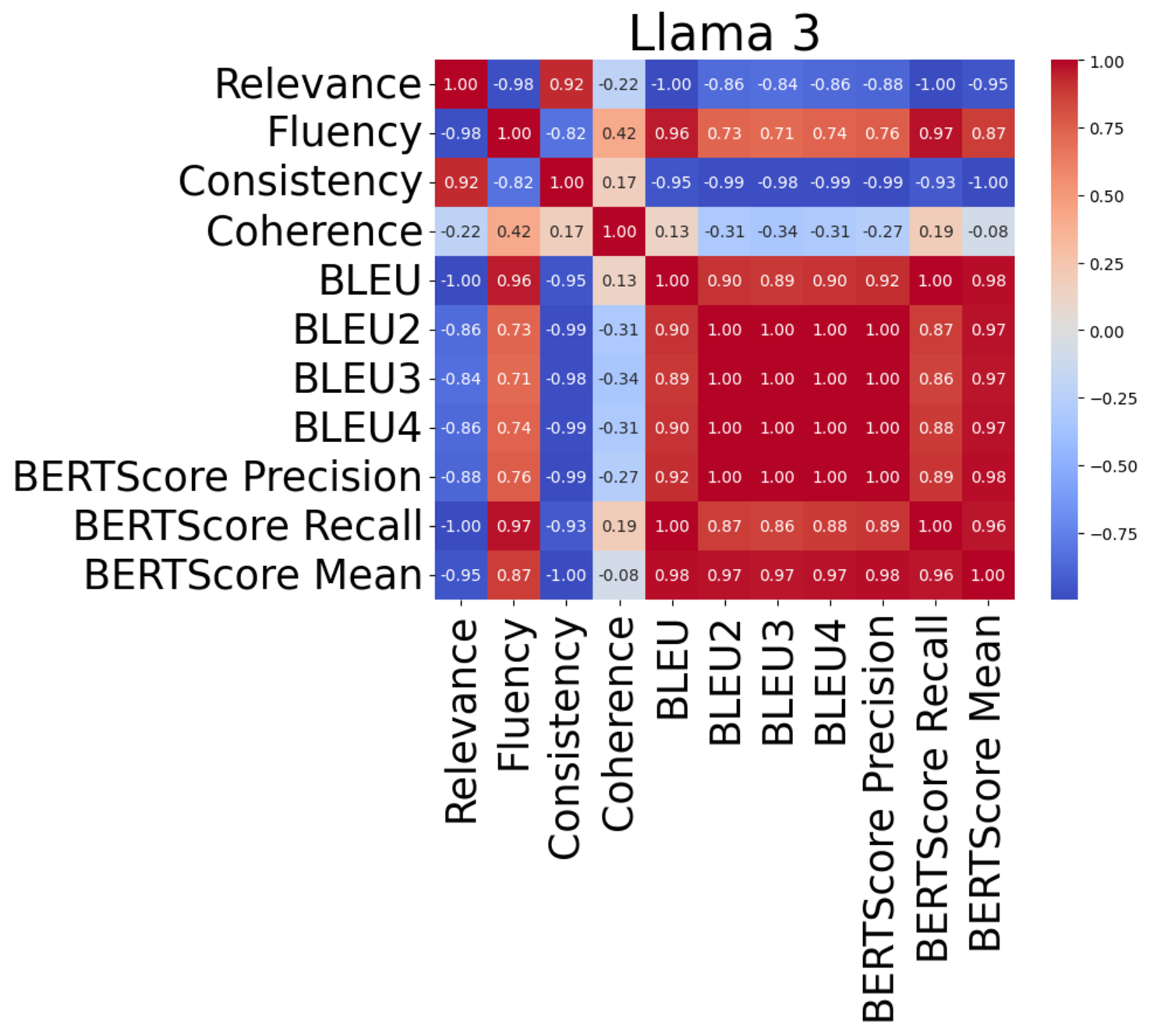}
        \caption*{WikiBio (Llama 3)}
    \end{subfigure}
    \hfill
    \begin{subfigure}[b]{0.45\textwidth}
        \centering
        \includegraphics[width=0.8\textwidth]{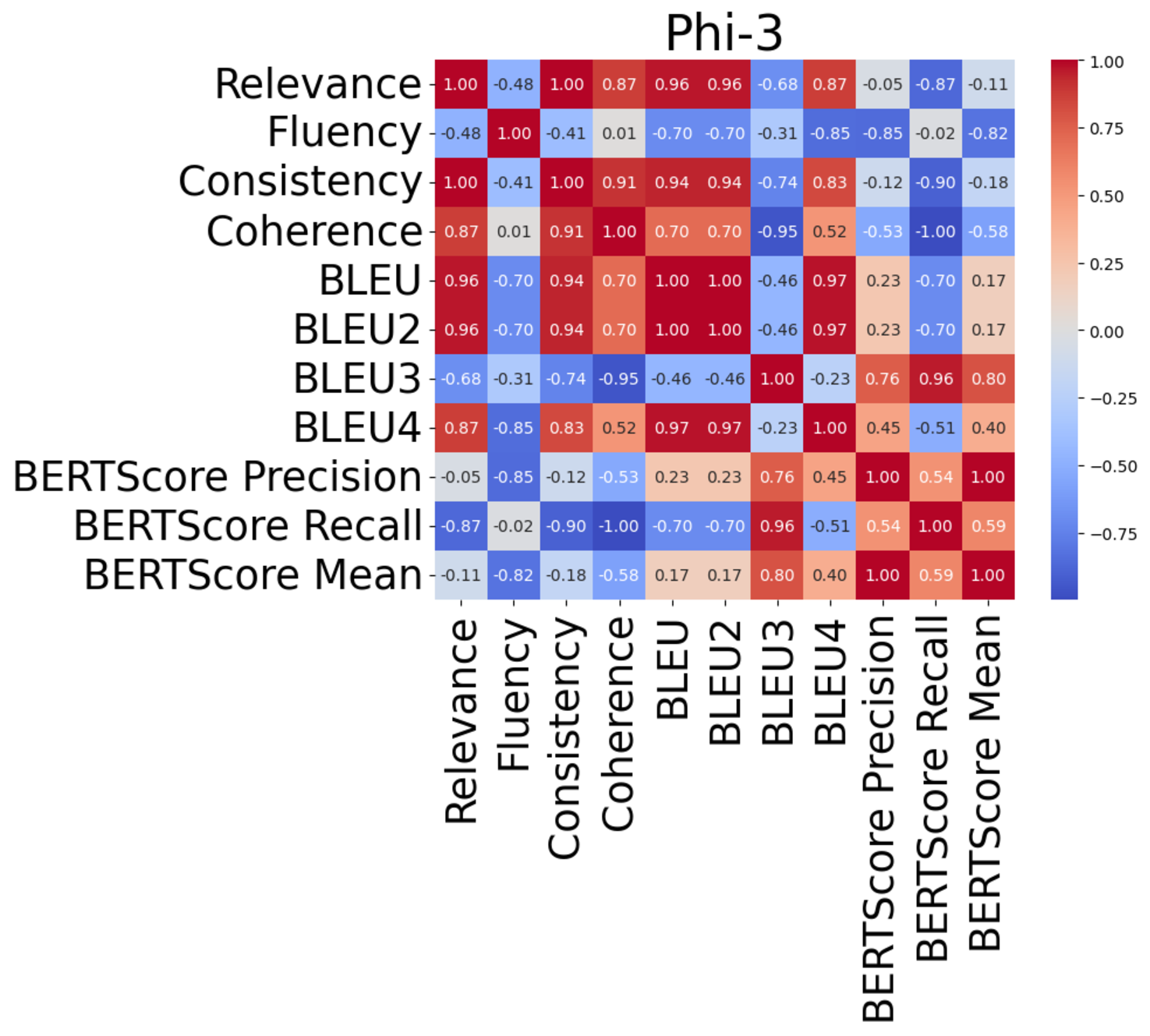}
        \caption*{WikiBio (Phi-3)}
    \end{subfigure}
    
    \vspace{0.5cm}
    
    \begin{subfigure}[b]{0.45\textwidth}
        \centering
        \includegraphics[width=0.8\textwidth]{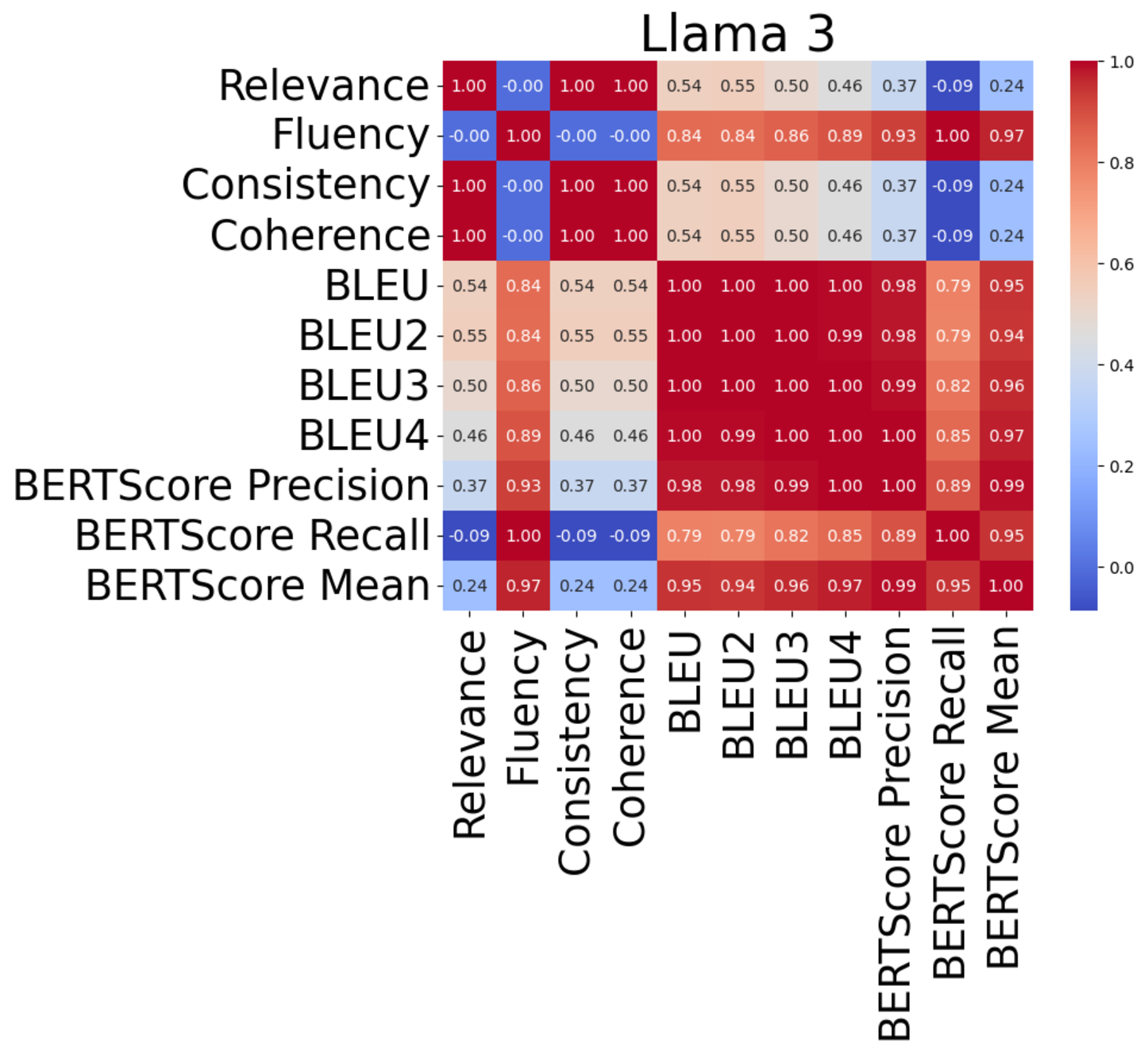}
        \caption*{Recent Wikipedia pages (Llama 3)}
    \end{subfigure}
    \hfill
    \begin{subfigure}[b]{0.45\textwidth}
        \centering
        \includegraphics[width=0.8\textwidth]{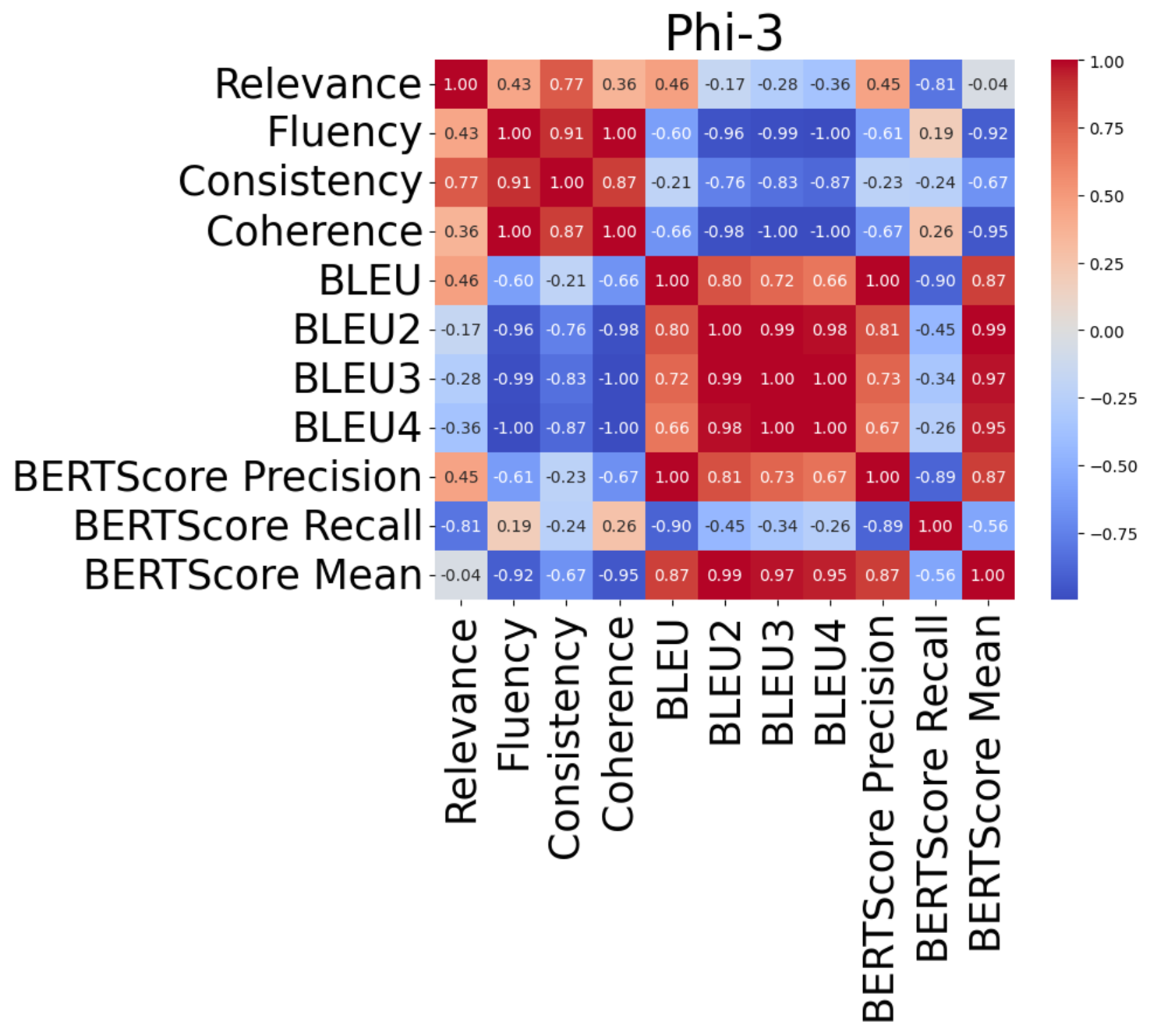}
        \caption*{Recent Wikipedia pages (Phi-3)}
    \end{subfigure}

    \caption{Correlation matrix comparisons of all metrics across different prompting strategies on the WikiBio and recent Wikipedia datasets using Llama 3 and Phi-3 models.}
    \label{fig:wiki-heatmap}
\end{figure}

Figure \ref{fig:wiki-heatmap} shows the correlation heatmap of all metrics across different prompting settings. It can be seen that there is a strong correlation between BLEU scores and BERTScores, indicating a consistent pattern, especially with the Llama 3 model.
However, while a closer correlation between LLM self-evaluation metrics and other metrics is observed with Llama 3, it does not exhibit a meaningful relationship, as it does not follow any consistent pattern. Additionally, the strong internal correlation between the self-evaluation metrics does not provide any significant insights, as the differences in scores were too small to be considered meaningful, according to Table \ref{tab:WikiBio_0-1-few-shot-comparision} and Table \ref{tab:recent_wiki_0-1-few-shot-comparision}. The self-evaluation scores mostly show the maximum values, indicating the LLM's tendency to prefer its own generated texts.

\begin{table}[t]
\centering
\caption{Comparison study of different prompting strategies on the recent Wikipedia pages.}
\label{tab:recent_wiki_0-1-few-shot-comparision}
\begin{tabular}{lccc|ccc}
\toprule
& \multicolumn{3}{c|}{Llama 3} & \multicolumn{3}{c}{Phi-3} \\
Metrics                        & Zero Shot & Single Shot & Few Shot & Zero Shot & Single Shot & Few Shot \\
\midrule
Relevance                  & \textbf{5.000} & 4.94 & \textbf{5.000} & 4.938    & \textbf{5.000} & 4.706 \\
Fluency                   & 2.875      & 2.938      & \textbf{3.000}    & 2.625     & 2.389    & 2.353 \\
Consistency               & \textbf{5.000}      & 4.750       & \textbf{5.000}    & \textbf{5.000}     & 4.889    & 4.765 \\
Coherence                 & \textbf{5.000}      & 4.938      & \textbf{5.000}    & \textbf{5.000}     & 4.944    & 4.941 \\
\hline
BLEU                     & 0.417   & 0.413    & \textbf{0.489} & 0.405  & \textbf{0.425}    & 0.410 \\
BLEU2                     & 0.367   & 0.363    & \textbf{0.429} & 0.355  & \textbf{0.385}    & 0.380 \\
BLEU3                     & 0.301   & 0.301    & \textbf{0.358} & 0.293  & \textbf{0.316}   & 0.315 \\
BLEU4                     & 0.250   & 0.253    & \textbf{0.305} & 0.244  & 0.262    & \textbf{0.263} \\
BERTScore-Precision            & 0.506   & 0.512    & \textbf{0.546} & 0.515  & \textbf{0.530}   & 0.519 \\
BERTScore-Recall               & 0.594   & 0.609    & \textbf{0.620} & 0.599  & 0.592    & \textbf{0.601} \\
BERTScore-Mean                 & 0.542   & 0.552    & \textbf{0.577} & 0.549  & \textbf{0.556}    & 0.554 \\
\bottomrule
\end{tabular}
\end{table}



\subsection{MaRDI}

Figure \ref{fig:mardi-dataset-example} presents a table data extracted from MaRDI Portal on the first row and the generated biography using few-shot prompting with Llama 3 on the second row. This example demonstrates the model's ability to accurately interpret the properties and values without generating hallucinations. The resulting biography is coherent, consistent, and easy to read and understand and covers the information provided in the table.

We manually observed that the few-shot approach, when provided with well-crafted demonstrative examples significantly reduced hallucinations in generated text. 
This reduction was evident not only when compared to the zero-shot approach but also when compared to the single-shot approach, where the model was only shown one pair of table and biography. Notably, Llama 3 demonstrated a lower incidence of hallucinations, particularly in cases of missing information. This finding underscores the importance of careful selection of demonstrative examples to cover similarity and diversity.
Figure \ref{fig:mardi_0-1-few-shot-comparision} illustrates an example of how different prompting strategies can lead to better outcomes when dealing with incomplete information. Compared to the example shown in Figure \ref{fig:mardi-dataset-example}, the structured data in this example contains only a few properties. In the zero-shot scenario, the model generated content that is not relevant to the provided table, leading to hallucinations. The single-shot approach mitigates these hallucinations, guided by a well-crafted example that sets a clear standard for what a good biography should look like. The few-shot approach, however, performed best, as it effectively avoids adding extraneous information. This success is likely due to one of the examples in the few-shot prompt including missing information, which helped the model learn to avoid generating unrelated content, which is highlighted in the last row of Figure  \ref{fig:mardi_0-1-few-shot-comparision}. However, we manually observed that Phi-3 was less responsive to these examples and continued to generate unrelated text in similar cases.

The improvement with Llama 3 based on our manual evaluation contrasts with the results presented in table  \ref{tab:mardi_evaluation}, which shows the LLM self-evaluation on the MaRDI dataset. This table shows that the differences between zero-shot, single-shot, and few-shot strategies across various metrics are relatively small.
We highlighted this contrast in second row of Figure \ref{fig:mardi_0-1-few-shot-comparision} for the corresponding MaRDI example in the first row. Despite the evident improvement in the generated output using the few-shot prompt using Llama 3, the LLM self-evaluation does not reflect these differences.
Based on the MaRDI evaluation and other results from the Wikipedia leading paragraph, it can be concluded that the LLM self-evaluation method still lacks the sensitivity needed to detect meaningful distinctions in performance. These findings suggest that, although LLM self-evaluation shows potential in other domains \cite{zhou2022large}, the current approaches may not yet be fully reliable for assessing table-to-text generation tasks.





\begin{table}[h!]
\centering
\caption{Comparison study of different prompting strategies on MaRDI with LLM self-evaluation.}
\label{tab:mardi_evaluation}
\begin{tabular}{lccc|ccc}
\toprule
Metrics      & \multicolumn{3}{c|}{Llama 3}    & \multicolumn{3}{c}{Phi-3}   \\
                       & Zero Shot & Single Shot & Few Shot & Zero Shot & Single Shot & Few Shot \\
\midrule
Relevance Score    & 4.827            & 4.820              & 4.860           & 4.900          & 4.600            & 4.800         \\
Fluency Score     & 3.000            & 2.973              & 2.747           & 2.980          & 2.943            & 2.947         \\
Consistency Score  & 4.436            & 4.547              & 5.000           & 4.613          & 4.147            & 4.493         \\
Coherence Score    & 5.000            & 4.947              & 4.527           & 4.980          & 4.886            & 4.927         \\
\bottomrule
\end{tabular}
\end{table}

\begin{figure}
\centering
\caption{An example of a MaRDI Person Profile with lacking information. The output shows how the few-shot approach helps reduce hallucination with Llama 3. This affects the Phi-3 less.}
\label{fig:mardi_0-1-few-shot-comparision}
\footnotesize
\begin{tabular}{l|l}
\textbf{Property} & \textbf{Value} \\ \hline
MaRDI profile type & MaRDI person profile \\ \hline
educated at & École nationale du génie de l'eau et de l'environnement de Strasbourg \\ \hline
given name & David \\ \hline
instance of & human \\ \hline
occupation & researcher \\ \hline
sex or gender & male \\
\hline
\end{tabular}
\vspace{0.2cm}

\centering
\vspace{0.5cm}

\textbf{Llama 3}\\
\vspace{0.3cm} 
\begin{minipage}{0.32\textwidth}
\begin{tcolorbox}[colback=gray!10, colframe=black, title={zero-shot prompting}, boxrule=0.5mm, fonttitle=\small, boxsep=1mm, left=1mm, right=1mm, top=1mm, bottom=1mm]
\footnotesize
David is a researcher who was educated at École nationale du génie de l'eau et de l'environnement de Strasbourg. With a strong foundation in his field, he has built a successful career as a researcher. Throughout his professional journey, David has made significant contributions to his area of expertise, earning recognition and respect from his peers. His academic background at École nationale du génie de l'eau et de l'environnement de Strasbourg has provided him with a solid understanding of the complexities in his field, enabling him to tackle complex research projects with ease.
\end{tcolorbox}
\end{minipage}%
\hfill
\begin{minipage}{0.32\textwidth}
\begin{tcolorbox}[colback=gray!10, colframe=black, title={single-shot prompting}, boxrule=0.5mm, fonttitle=\small, boxsep=1mm, left=1mm, right=1mm, top=1mm, bottom=1mm]
\footnotesize
David was educated at the prestigious École nationale du génie de l'eau et de l'environnement de Strasbourg. As a researcher, he has dedicated his career to advancing knowledge in his field. With a strong academic foundation, David has made significant contributions to his area of expertise, solidifying his position as a respected expert.
\end{tcolorbox}

\begin{tcolorbox}[colback=gray!10, colframe=black, title={few-shot prompting}, boxrule=0.5mm, fonttitle=\small, boxsep=1mm, left=1mm, right=1mm, top=1mm, bottom=1mm]
\footnotesize
David is a researcher who holds a degree from École nationale du génie de l'eau et de l'environnement de Strasbourg.
\end{tcolorbox}
\end{minipage}%
\hfill
\begin{minipage}{0.32\textwidth}
\begin{tikzpicture}
    \begin{axis}[
        ybar,
        bar width=0.2cm,
        width=6cm,
        height=4cm,
        xlabel={Evaluation Metrics for Llama 3 performance},
        ylabel={Scores},
        symbolic x coords={Relevance, Fluency, Consistency, Coherence},
        xtick=data,
        xticklabel style={rotate=45, anchor=east},  
        ymin=0, ymax=6,
        enlarge x limits={abs=0.5cm},
        legend style={at={(0.5,-1)},
        anchor=north,legend columns=-1},
        nodes near coords,
        nodes near coords align={vertical},
        every node near coord/.append style={font=\footnotesize},
    ]
    
    \addplot[
        fill={rgb,255:red,150; green,150; blue,200},
    ] coordinates {(Relevance,5) (Fluency,3) (Consistency,5) (Coherence,5)};
    
    \addplot[
        fill={rgb,255:red,200; green,150; blue,150},
    ] coordinates {(Relevance,5) (Fluency,3) (Consistency,5) (Coherence,5)};
    
    \addplot[
        fill={rgb,255:red,150; green,200; blue,150},
    ] coordinates {(Relevance,5) (Fluency,2) (Consistency,5) (Coherence,5)};
    
    \legend{Zero-shot, Single-shot, Few-shot}
    
    \end{axis}
\end{tikzpicture}

\end{minipage}

\centering 
\vspace{0.3cm}

    
    
    
    
    

\vspace{0.2cm}
\centering
\textbf{Phi-3} \\
\vspace{0.3cm}
\begin{minipage}{0.32\textwidth}
\begin{tcolorbox}[colback=gray!10, colframe=black, title={zero-shot prompting}, boxrule=0.5mm, fonttitle=\small, boxsep=1mm, left=1mm, right=1mm, top=1mm, bottom=1mm]
\footnotesize
David is renowned for his dedication to research in environmental engineering. His academic journey began at École nationale du génie de l'eau et de l'environnement de Strasbourg, where he gained a solid foundation in water and environment management techniques. Over the years, David has cultivated extensive expertise within the field, contributing valuable insights that continue to shape our understanding of ecological sustainability.
\end{tcolorbox}
\end{minipage}%
\hfill
\begin{minipage}{0.32\textwidth}
\begin{tcolorbox}[colback=gray!10, colframe=black, title={single-shot prompting}, boxrule=0.5mm, fonttitle=\small, boxsep=1mm, left=1mm, right=1mm, top=1mm, bottom=1mm]
\footnotesize
David pursued his academic journey at École nationale du génie de l'eau et de l'environnement de Strasbourg, where he honed his expertise in water and environmental engineering. As a dedicated researcher in the field, 
David has contributed valuable knowledge to the scientific community through rigorous study and exploration of innovative solutions for managing water resources sustainably.
\end{tcolorbox}

\end{minipage}%
\hfill
\begin{minipage}{0.32\textwidth}
\begin{tcolorbox}[colback=gray!10, colframe=black, title={few-shot prompting}, boxrule=0.5mm, fonttitle=\small, boxsep=1mm, left=1mm, right=1mm, top=1mm, bottom=1mm]
\footnotesize
David is a dedicated researcher who pursued his studies at École nationale du génie de l'eau et de l'environnement de Strasbourg. With an extensive background in environmental engineering, he contributes valuable insights to the field of sustainable development and water management practices through ongoing research projects.
\end{tcolorbox}
\end{minipage}

\end{figure}

\subsection{ToTTo}

\begin{figure}[h]
    \centering
    \begin{minipage}{0.45\textwidth}
        \centering
        \begin{tabular}{p{1cm}p{0.7cm}p{0.7cm}p{0.8cm}|p{0.7cm}p{0.7cm}p{0.7cm}}
            \toprule
            & \multicolumn{3}{c}{Llama 3} & \multicolumn{3}{c}{Phi-3} \\
            Metric  & Zero-shot & Single-shot & Few-shot & Zero-shot & Single-shot & Few-shot  \\ 
            \midrule
            BLEU   & 0.585  & 0.545  & \textbf{0.635}  & 0.560  & \textbf{0.610}  & 0.601  \\
            BLEU2   & 0.541  & 0.506  & \textbf{0.597}  & 0.513  & \textbf{0.576}  & 0.554  \\
            BLEU3   & 0.487  & 0.464  & \textbf{0.551}  & 0.461  & \textbf{0.519}  & 0.498  \\
            BLEU4   & 0.446  & 0.431  & \textbf{0.514}  & 0.421  & \textbf{0.476}  & 0.456  \\
            \bottomrule
        \end{tabular}
        \caption{Comparison study of different prompting strategies on the ToTTo Dataset.}
        \label{tab:ToTTo-0-1-few-shot-comparision}
    \end{minipage}%
    \hfill
    \begin{minipage}{0.5\textwidth}
        \centering
        \begin{tikzpicture}
            \begin{axis}[
                ybar,
                bar width=0.2cm,
                width=6cm,
                height=5cm,
                xlabel={Models},
                ylabel={Summed Overall Scores},
                symbolic x coords={Llama 3, Phi-3},
                xtick=data,
                ymin=0, ymax=6,
                xtick align=center,
                xtick style={draw=none},
                enlarge x limits=0.3,  
                legend style={at={(0.5,-0.15)},
                anchor=north,legend columns=-1},
                nodes near coords,
                nodes near coords align={vertical},
                every node near coord/.append style={font=\footnotesize},
                ylabel near ticks,
            ]
            \addplot[
                ybar,
                bar shift=-0.15cm,
                fill={rgb,255:red,200; green,200; blue,100},
                draw=black,
            ] coordinates {(Llama 3,4.8539) (Phi-3,4.4738)};
        
            \addplot[
                ybar,
                bar shift=0.15cm,
                fill={rgb,255:red,100; green,200; blue,200},
                draw=black,
            ] coordinates {(Llama 3,4.230956) (Phi-3,3.4397)};
        
            \legend{GPT-modified prompt, human Prompts}
        
            \end{axis}
        \end{tikzpicture}
        \caption{Effect of Human-Crafted Prompts Versus GPT-4-Generated Prompts.}
        \label{fig:humancrafted-vs-GPT4generated-prompt}
    \end{minipage}
\end{figure}

Table \ref{tab:ToTTo-0-1-few-shot-comparision} demonstrates the comparison study of the zero-shot, single-shot, and few-shot prompting on the ToTTo dataset using  Llama 3 and Phi-3 models. 
The results indicate that Llama 3 benefits more from the few-shot approach compared to Phi-3, which aligns with findings from other experiments. The reduction in Phi-3's performance with the few-shot method, as opposed to single-shot, mirrors a similar trend observed in recent Wikipedia pages (Table \ref{tab:recent_wiki_0-1-few-shot-comparision}) as well. This reduction can likely be attributed Phi-3's sensitivity to context length. Including multiple examples in the prompt might create a context that is too lengthy for Phi-3, as an SLM, to effectively process and understand, although it has the same context length as Llama 3.
Our manual evaluation of 10 randomly selected samples further supports this, as the samples generated using the few-shot approach with Llama 3 showed noticeable improvements in  the quality.
We mentioned earlier that we modified the human-crafted prompts for the language models using GPT-4, and for ToTTo dataset we made an attempt to compare the original prompt with the modified prompt in Figure \ref{fig:humancrafted-vs-GPT4generated-prompt}. The results indicate that the performance of both Llama 3 and Phi-3 models improved with GPT-4-modified prompts compared to the human-crafted ones.


\section{Related Work}
The notable advancements in LMs \cite {touvron2023llama, achiam2023gpt, abdin2024phi, team2023gemini} have driven a major shift in the fields of controllable text generation and data interpretation. Leveraging these developments, recent research has explored table processing across various scenarios, including question answering \cite{cheng2022binding}, fact checking \cite{das2023state}, and real-world information seeking \cite{zhao2023investigating}.
While fine-tuning models \cite{chen2020logical, liu2022plog, zhao2023loft} on specific tasks remains popular, it often requires substantial amounts of high-quality data. Moreover, these table-based tasks encompass a wide variety of input-output formats and domains, presenting significant variability across task types. Studies such as UnifiedSKG \cite{xie2022unifiedskg} have aimed to standardize diverse table-based tasks by converting them into a unified text-to-text format. 
Instruction-tuning is another promising approach, as explored by models like TableLlama \cite{zhang2023tablellama} and Table-GPT \cite{li2023table}. This approach involves constructing instruction tuning datasets and continuing the pretraining LMs, such as Llama 2 (7B) \cite{touvron2023llama} in TableLlama and GPT-3.5 in Table-GPT. 
While instruction-tuning has demonstrated promising results and enhanced generalization capabilities, the process of curating new datasets and pretraining large models comes with significant computational and resource costs. Moreover, instruction-tuning does not consistently outperform the fine-tuning of smaller, pre-trained language models in terms of accuracy. Previous research \cite{li2024dawn} suggests that across various metrics and domains, there is no definitive advantage of LLM-based solutions over pre-LLM approaches.
Although Table-GPT demonstrates superior performance on table-related tasks compared to GPT-3.5 and ChatGPT through continued training on GPT-3.5, the associated training costs remain a prohibitive factor for many enterprises looking to deploy models privately.
Despite these efforts, recent findings \cite{chen2022large, lu2024large} indicate that LMs, through in-context learning, can outperform fine-tuned models on table-processing tasks. In contrast to fine-tuning, in-context learning (1) requires only a few annotations or demonstrations as prompts, and (2) conducts inference without modifying the model parameters, reducing the need for extensive fine-tuning. 
The effectiveness of in-context learning largely depends on how well demonstrative examples are selected and organized within the prompt. For instance, Nan et al. \cite{nan2023enhancing-} suggest that to improve the text-to-SQL capabilities of LMs, it is important to consider not only similarity but also diversity among the examples. 
We explored this potential in our MaRDI use case.
In-context learning approaches heavily studied the superiority of GPT families (GPT3, 3.5, 4 and ChatGPT) \cite{chen2022large,zhao2023investigating} for various table processing tasks. We based our experiment on open-source models that can be run locally, alleviating concerns around privacy and the high costs. Additionally, we specifically targeted the table-to-text generation task, which remains surprisingly underexplored in the current literature.
Another factor of the effectiveness of in-context learning is the unpredictability of plain language prompts.
Plain language prompts may fail to produce desired results, requiring users to experiment with various instructions \cite{sanh2021multitask, wei2021finetuned}. This is due to the lack of transparency in how LMs process the instructions for the human. We modified human-crafted prompts using stronger LMs like GPT-4 to instruct other LMs and explored its potential in ToTTo data analysis to improve the outcomes. This approach does not raise concerns about data privacy or higher costs, as the prompts are only used once and solely for describing the task to the other LMs.

\section{Discussion \& Conclusion}
In this study, we employed a comprehensive prompting strategy across multiple datasets to evaluate two language models, an LLM and an SLM, for table-to-text generation tasks. Our aim was to assess the models' effectiveness in both standard and real-world applications, including the WikiBio, ToTTo datasets, and our use case MaRDI data. This study suggests that language models can benefit from in-context learning by providing a set of proper examples to the model for table-to-text generation.

We investigated the potential influence of prior data exposure using the WikiBio dataset. To address this, we retrieved recent Wikipedia pages and formatted them similarly to the WikiBio dataset, then conducted experiments on this newly retrieved data. The key finding from this study (Tables \ref{tab:WikiBio_0-1-few-shot-comparision} and \ref{tab:recent_wiki_0-1-few-shot-comparision}) was that we observed a more significant improvement in biography generation on the recent Wikipedia pages compared to the WikiBio dataset. We attribute this effect to the likelihood that the language models had not encountered the recent data during training, making the impact of providing examples in the prompts more pronounced. This conclusion was supported by both automated metrics and manual evaluations. 

We compared the performance of Llama 3, as an LLM, with Phi-3, an SLM. As expected, Llama 3 generally outperformed Phi-3 across various settings and datasets, reflecting the advantages of larger models in table-to-text tasks. However, the primary goal of this comparison was to analyze how both models responded to few-shot prompting. While both improved with the inclusion of examples, Llama 3 showed consistently greater benefits from additional examples. Notably, in our experiments with the MaRDI dataset, a real-world application, Llama 3 demonstrated a significant reduction in hallucinations, an improvement not observed with Phi-3 (Figure \ref{fig:mardi_0-1-few-shot-comparision}).

We studied the potential of LLM self-evaluation with CoT reasoning for measuring the table-to-text generation task using Llama 3, a powerful open-source LLM. Specifically, we aimed to assess the correlation of this metric, which does not require a ground truth for comparison, with BERTScore, which is known to align well with human judgment.
Interestingly, despite the potential of this strategy in other tasks \cite{liu2023g}, our study revealed that the self-evaluation strategy showed weak or no meaningful correlation with BERTScore (Figure \ref{fig:wiki-heatmap}). To investigate further, we analyzed the MaRDI dataset and manually observed a significant reduction in hallucinations when comparing the few-shot approach to the zero-shot method. However, despite this noticeable improvement, the LLM self-evaluation metrics showed minimal variation and failed to reflect the relevance between the provided table and the corresponding generated biography (Figure \ref{fig:mardi_0-1-few-shot-comparision}).
These results suggest that the current state-of-the-art LLM self-evaluation strategy, even with advanced open-source models like Llama 3, is not yet reliable for measuring table-to-text generation tasks. This can be explained by either the LLM favors their own generated outputs or failing to establish a meaningful connection between the table and the generated text. In future work, we will explore providing observatory examples for evaluation model that might improve the alignment between LLM self-evaluation and human judgment.

Table processing studies have often demonstrated LM performance on benchmark datasets that are known to be well-structured.  However, current public datasets, such as WikiBio and ToTTo, commonly used for benchmarking, have inherent limitations for table-to-text generation. For example, Wikipedia-based datasets like WikiBio often contain inconsistencies between infoboxes and the associated text. Similarly, the ToTTo dataset presents challenges where descriptions sometimes focus solely on a specific table cell, while in other cases, additional information from the table is included. This mismatch complicates accurate model evaluation.
This is where our study of the MaRDI use case, representing a real-world application, offers valuable insights. It built a foundation on how to design prompts and how to prepare examples where there is no ground truth in real-world table-to-text generation scenarios (section \ref{subsubsection:MaRDI}).

\bibliographystyle{plain}
\bibliography{references}




\end{document}